\documentclass[nohyperref]{article}

\usepackage{microtype}
\usepackage{graphicx}
\usepackage{subfigure}
\usepackage{booktabs} %

\usepackage{hyperref}
\usepackage{paralist}

\makeatletter
\def\url@leostyle{%
  \@ifundefined{selectfont}{\def\UrlFont{}}%
  {\def\UrlFont{}}%
}
\makeatother
\usepackage[accepted]{icml2022}

\usepackage{amsmath}
\usepackage{amssymb}
\usepackage{mathtools}
\usepackage{amsthm}

\usepackage[capitalize,noabbrev]{cleveref}

\theoremstyle{plain}

\theoremstyle{definition}

\theoremstyle{remark}

\usepackage[textsize=tiny]{todonotes}

\usepackage{bbold}
\usepackage{color}
\usepackage[small,bf]{caption}

\usepackage{graphicx} %
	\setkeys{Gin}{width=\textwidth, height=\textheight, keepaspectratio}
\newif\ifcomment
\commenttrue

\ifcomment
	\newcommand{\sof}[1]{\textbf{\em\color{red}SOF: #1}}
	\newcommand{\kai}[1]{\textbf{\em\color{green}KAI: #1}}
	\newcommand{\gvg}[1]{\textbf{\em\color{blue}GG: #1}}
\else
		\newcommand\sof[1]{}
		\newcommand\kai[1]{}
		\newcommand\gvg[1]{}
\fi
\newcommand{\descr}[1]{\smallskip\noindent\textbf{#1}}

 \definecolor{linkcol}{RGB}{0,70,25}
 \definecolor{citecol}{RGB}{0,70,25}
 \definecolor{urlcol}{RGB}{70,0,25}

\usepackage{listings}
\icmltitlerunning{dpart: Differentially Private Autoregressive Tabular}

\begin{document}

\twocolumn[
\icmltitle{\emph{dpart}: Differentially Private Autoregressive Tabular,\\ a General Framework for Synthetic Data Generation}

\icmlsetsymbol{equal}{*}

\begin{icmlauthorlist}
\icmlauthor{Sofiane Mahiou}{hazy}
\icmlauthor{Kai Xu}{hazy,ed}
\icmlauthor{Georgi Ganev}{hazy,ucl}
\end{icmlauthorlist}

\icmlaffiliation{hazy}{Hazy, London, UK}
\icmlaffiliation{ucl}{University College London, London, UK}
\icmlaffiliation{ed}{University of Edinburgh, Edinburgh, UK}

\icmlcorrespondingauthor{Sofiane Mahiou}{sofiane@hazy.com}

\icmlkeywords{Differential Privacy, Synthetic Data, Open Source}

\vskip 0.3in
]

\printAffiliationsAndNotice{} %

\begin{abstract}
We propose a general, flexible, and scalable framework \emph{dpart}, an open source Python library for differentially private synthetic data generation.
Central to the approach is autoregressive modelling---breaking the joint data distribution to a sequence of lower-dimensional conditional distributions, captured by various methods such as machine learning models (logistic/linear regression, decision trees, etc.), simple histogram counts, or custom techniques.
The library has been created with a view to serve as a quick and accessible baseline as well as to accommodate a wide audience of users, from those making their first steps in synthetic data generation, to more experienced ones with domain expertise who can configure different aspects of the modelling and contribute new methods/mechanisms.
Specific instances of \emph{dpart} include Independent, an optimized version of PrivBayes, and a newly proposed model, dp-synthpop.

Code: \url{https://github.com/hazy/dpart}.
\end{abstract}

\section{Introduction}
Sharing data across and within  enterprises is essential for numerous innovation and commercial initiatives but is restricted by privacy and ethical concerns.
Privacy-preserving synthetic data, relying on rigorous privacy definitions such as Differential Privacy~(DP) and machine learning generative models, is a promising solution as it could serve a drop-in replacement of the real data while maintaining the individuals' privacy.
It has been enjoying a great deal of interest by the research community~\cite{jordon2022synthetic} as well as government organizations such as the NHS~\cite{nhs21ae}, US Census Bureau~\cite{benedetto2018creation}, and NIST~\cite{nist2018differential, nist2020differential}.

While there is a rich literature with DP models for synthetic data generation~\cite{acs2018differentially, xie2018differentially, jordon2018pate, vietri2020new, zhang2021privsyn, mckenna2021winning, aydore2021differentially, liu2021iterative},\footnote{For a more comprehensive list with DP generative models see: \url{https://github.com/ganevgv/dp-generative-models}} most of them are slow to run, require considerable computational resources, not accessible to users with different levels of expertise, not flexible enough, and could be hard coded to a specific problem.
To fill this gap, we propose \emph{D}ifferetially \emph{P}rivate \emph{A}uto\emph{R}egressive \emph{T}abular, or \emph{dpart} -- an open source library (MIT license) for efficient DP synthetic data generation.
While, it is not meant to achieve state-of-the-art results, \emph{dpart} is easy to use and is highly customizable/configurable, allowing users to quickly experiment and achieve competitive baseline results.

In terms of interface and under-the-hood modelling, we were inspired by diffprivlib~\cite{holohan2019diffprivlib} (open source DP models and mechanisms), sklearn~\cite{sklearn2011pedregosa} (API), DataSynthesizer~\cite{ping2017datasynthesizer} (graphical model), and synthpop~\cite{nowok06synthpop} (autoregressive model and input arguments).
\emph{dpart} is meant to be utilized by both non-experts---one could fit a model in a single line of code, \emph{dpart().fit(X)} (assuming the model is imported and the data is loaded)---and by more sophisticated users who wish to input domain knowledge (e.g., the underlying data dependency graph or the distribution of the privacy budget across modeling steps or columns) or contribute their custom methods.

We present an overview of \emph{dpart} in Sec.\ref{sec:2}, three specific instances (\emph{Independent}, \emph{PrivBayes}, and \emph{dp-synthpop}) in Sec.\ref{sec:3} and compare their performance in Sec.\ref{sec:4}.

\descr{Main Contributions:}
\begin{compactitem}
  \item Implement and open source a general and flexible framework for DP synthetic data generation, based on autoregressive modelling, to serve as a quick and efficient baseline.
  \item As a specific use case of \emph{dpart}, we modify and improve the speed of \emph{PrivBayes}~\cite{zhang2017privbayes} by 20x.
  \item As another use case, we propose a DP version of synthpop~\cite{nowok06synthpop}, which we name \emph{dp-synthpop}.
\end{compactitem}

\begin{figure}[t!]
	\centering
	\subfigure{\includegraphics[width=0.35\textwidth]{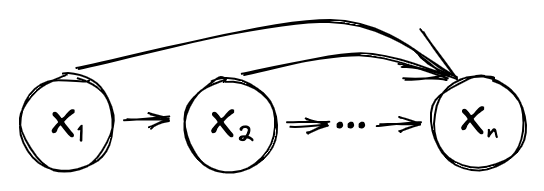}}
	\caption{\emph{dpart} framework.}
	\label{fig:1}
\end{figure}

\section{Overview}
\label{sec:2}

\descr{Framework summary.}

\emph{dpart} is a general and flexible framework for building an effective DP generative model.
The overall training flow relies on autoregressive generative model (Fig.~\ref{fig:1}) and could be broken down to two main steps.
First, identifying/specifying a visit order or a prediction matrix that describes how the joint distribution is broken down to a series of lower-dimensional conditionals (the dependency).
In other words, if the dataset $D$ is a collection of $k$-dimensional datapoints $\boldsymbol{x}$ then:

$P(\boldsymbol{x}) = \prod_{i=1}^{k} P(x_{i}|x_1,x_2,...,x_{i-1}) = \prod_{i=1}^{k} P(x_{i}|\boldsymbol{x}_{<i})$

Second, given the series of conditionals, they are sequentially estimated by fitting predictive models (sampler methods).
To generate synthetic data, the fitted sampler methods are used to generate one column at a time.

\descr{Installation.}

\emph{dpart} is written in Python due to the language popularity among data scientists as well as machine learning researchers and practitioners.
It can be installed using pip:

\begin{lstlisting}[language=Python,basicstyle=\tiny,frame=tb]
>>> pip install dpart
\end{lstlisting}

\descr{Training \& Generation.}

We look at the arguments responsible for 1) building/specifying the dependency, 2) the methods for estimating the conditional distributions, and 3) privacy budget distribution.

1) dependency arguments:
\begin{compactitem}
  \item \emph{visit\_order}: a list representing the order in which the the joint distribution is broken down into a sequence of conditionals.
  \item \emph{prediction\_matrix}: a dictionary specifying the collection of all (already visited) columns to be used as features/predictors for each unvisited column.
  If specified, the visit order is identified through khan sorting.
  \item Alternatively, \emph{prediction\_matrix} could be set to ``infer.''
  In this case, an optimal network, maximizing the mutual information between the columns, is built.
  Furthermore, in order to reduce computational complexity, \emph{n\_parents}, a maximum number of columns to be considered as features, could be specified.
  DP is guaranteed through the Exponential mechanism.
\end{compactitem}

At most one of \emph{visit\_order} and \emph{prediction\_matrix} could be used, as the two arguments conflict with each other.

2) methods arguments:
\begin{compactitem}
  \item \emph{methods}: a dictionary specifying the method each column should be modelled by.
  Columns must match the data type support of the selected method.
  A list with currently available methods is shown in Tab.~\ref{tab:1} and further explanations provided below.
\end{compactitem}

\begin{table}[t!]
  \centering
  \begin{tabular}[width=0.99\textwidth]{| c | c |}
    \hline
    Data type support   &   Method \\
    \hline
    \hline
    numerical only  & DP linear regression \\
    \hline
    categorical only  & DP logistic regression, \\
                      & DP decision tree, \\
                      & DP random forest \\
    \hline
    both   & DP conditional distribution, \\
           & DP histogram sample \\
    \hline
  \end{tabular}
  \caption{Methods.}
  \label{tab:1}
\end{table}

The methods can be split into the following three categories.

\emph{Numerical only} methods.
These methods can be applied on target columns with numerical data type (i.e., float, integer, datetime, and timedelta):
\begin{compactitem}
  \item regression method: it relies on fitting a DP regression model to predict the target column.
  In order to allow for non-deterministic behavior, the standard deviation of the residuals is captured in a DP way using the Laplace mechanism.
  During generation, new values are sampled by adding appropriate noise to the prediction from the trained regression.
  Currently available regression methods are: \emph{DP linear regression}.
\end{compactitem}

\emph{Categorical only} methods.
The methods below can be used on categorical columns with either an object, category, or boolean data type:
\begin{compactitem}
  \item classifier method: it fits a DP classification model that can output a conditional distribution.
  The available classification methods are: \emph{DP logistic regression, DP decision tree, and DP random forest.}
\end{compactitem}

\emph{Both numerical and categorical} methods:
\begin{compactitem}
  \item \emph{DP conditional distribution}: it captures and samples from a discretized joint distribution.
  Numerical data is binned using uniform binning to allow for a discrete representation and DP is satisfied by adding a Laplace noise to the counts before converting to a distribution.
  \item \emph{DP histogram sampler}: this method captures the marginal distribution of the target column without taking into account any input features.
  It is a specific use case of \emph{DP conditional distribution}.
\end{compactitem}

3) privacy budget arguments:
\begin{compactitem}
  \item \emph{epsilon}: a positive real number which defines the overall privacy budget to be used across the fitting step.
  \item Alternatively, a dictionary describing how the privacy budget can be split between the dependency and the methods steps could be provided.
  Furthermore, the user can break down the privacy budget between the methods for each column.
  \item \emph{bounds}: a dictionary specifying the range (minimum and maximum) for all numerical columns as well as the distinct categories for categorical columns.
  This prevents further privacy leakage.
  Alternatively, \emph{PrivacyLeakWarning} is displayed (see below).
\end{compactitem}

\descr{Troubleshooting.}

Inspired by diffprivlib, we adopt specific privacy (and other) warnings messages:
\begin{compactitem}
  \item \emph{PrivacyLeakWarning}: this warning is raised when privacy related input is missing.
  A good example is \emph{bounds} which must be provided to ensure that no further privacy leakage is incurred.
  However, if the \emph{bounds} are not provided, the algorithm will run and infer the missing bound values but will raise a warning (if epsilon has been provided).
  \item \emph{UserWarning}: this warning is raised when a method is not explicitly specified for a given column. The warning displays the default method which is used.
\end{compactitem}

\section{Specific Instances}
\label{sec:3}

While \emph{dpart} allows for a wide range of flexibility and customization, we configure and present three specific instances of the framework, which could easily be imported with a single line of code.
They are also summarized in Tab.~\ref{tab:2}.

\begin{table}[t!]
  \centering
  \begin{tabular}[width=0.99\textwidth]{| c | c |}
    \hline
    Model   &   Notes \\
    \hline
    \hline
    \emph{Independent} & simple baseline model \\
    & \cite{ping2017datasynthesizer, tao2021benchmarking} \\
    & \cite{stadler2022synthetic} \\
    \hline
    \emph{PrivBayes}  & optimized model \\
    & \cite{zhang2017privbayes, ping2017datasynthesizer} \\
    \hline
    \emph{dp-synthpop}  & new model, DP version of \\
    & \cite{nowok06synthpop} \\
    \hline
  \end{tabular}
  \caption{Specific instances of \emph{dpart}.}
  \label{tab:2}
\end{table}

\descr{Independent.}

\begin{figure}[t!]
	\centering
	\subfigure[\emph{Independent}]{\includegraphics[width=0.35\textwidth]{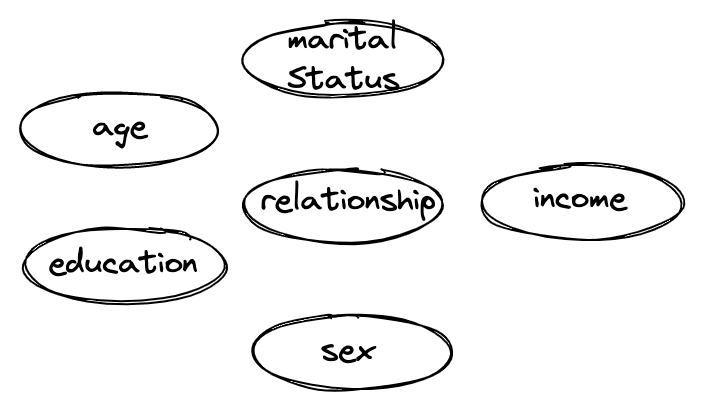}\label{fig:2a}}
	\subfigure[\emph{PrivBayes}, with $n\_parents=2$]{\includegraphics[width=0.35\textwidth]{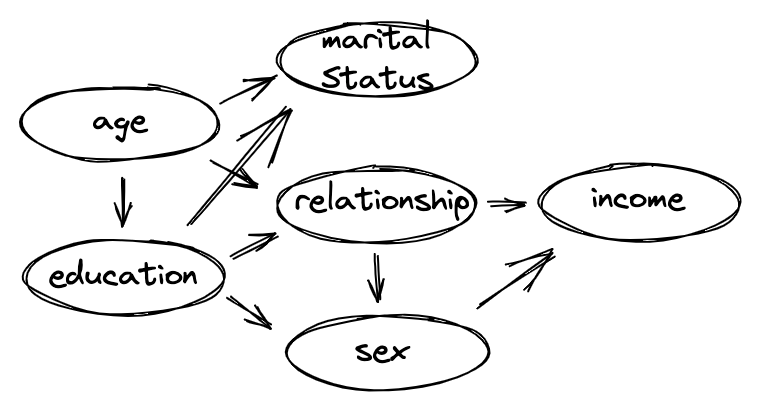}\label{fig:2b}}
	\subfigure[\emph{dp-synthpop}]{\includegraphics[width=0.35\textwidth]{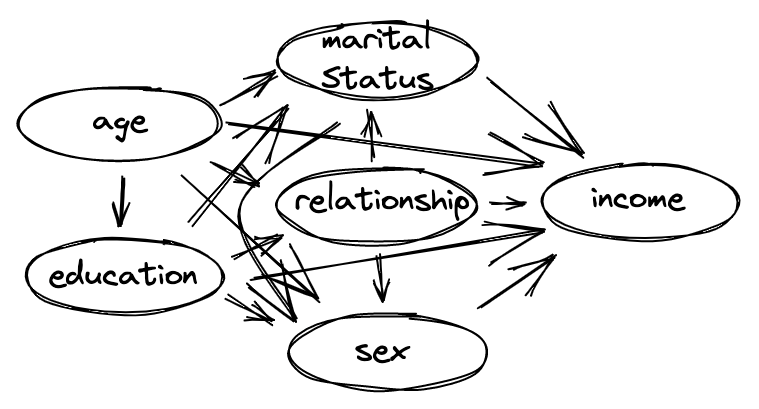}\label{fig:2c}}
	\caption{Dependencies of three specific instances of \emph{dpart} on simplified Adult.}
	\label{fig:2}
\end{figure}

This specific use case models all columns independently by using \emph{DP histogram sampler}.
The model has also been used as a baseline by~\cite{tao2021benchmarking, stadler2022synthetic} and while it looks very simple and naive, it has been shown that it could perform better than far more sophisticated models.
The dependency graph is presented in Fig.~\ref{fig:2a}.
The code excerpt below demonstrates how one could initiate, fit \emph{Independent}, and generate 1,000 rows for given privacy budget, dataset, and dataset bounds:

\begin{lstlisting}[language=Python,basicstyle=\tiny,frame=tb]
>>> from dpart.engines import Independent

>>> dpart_ind = Independent(epsilon, bounds=X_bounds).fit(X)
>>> synth_df = dpart_ind.generate(1000)
\end{lstlisting}

\descr{PrivBayes.}

\emph{PrivBayes} could also be seen as a sub case of \emph{dpart}.
We speed up the implementation offered by~\cite{ping2017datasynthesizer} by 20x by re-implementing the dependency-inference step. Further performance improvements could be achieved by proposing alternative, more efficient dependency-inference approaches.
A possible dependency graph produced by \emph{PrivBayes} is shown in Fig.~\ref{fig:2b}, while a code example could be found below:

\begin{lstlisting}[language=Python,basicstyle=\tiny,frame=tb]
>>> from dpart.engines import PrivBayes

>>> dpart_pb = PrivBayes(epsilon, bounds=X_bounds).fit(X)
>>> synth_df = dpart_pb.generate(1000)
\end{lstlisting}

\begin{figure*}[ht!]
	\centering
	\subfigure[Similarity]{\includegraphics[width=0.49\textwidth]{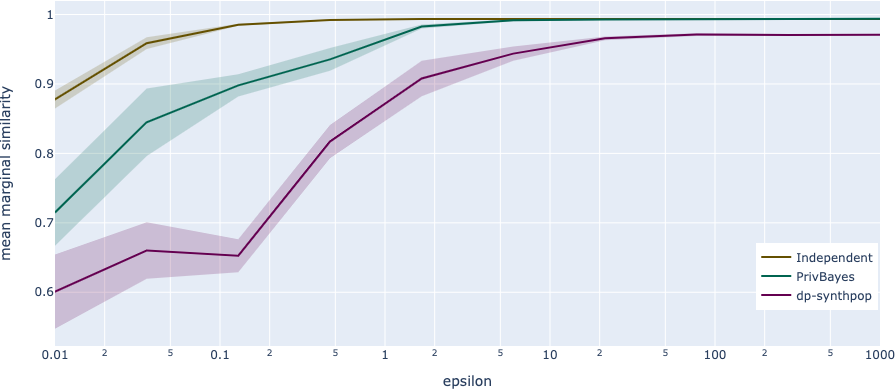}\label{fig:3a}}
	\subfigure[Accuracy]{\includegraphics[width=0.49\textwidth]{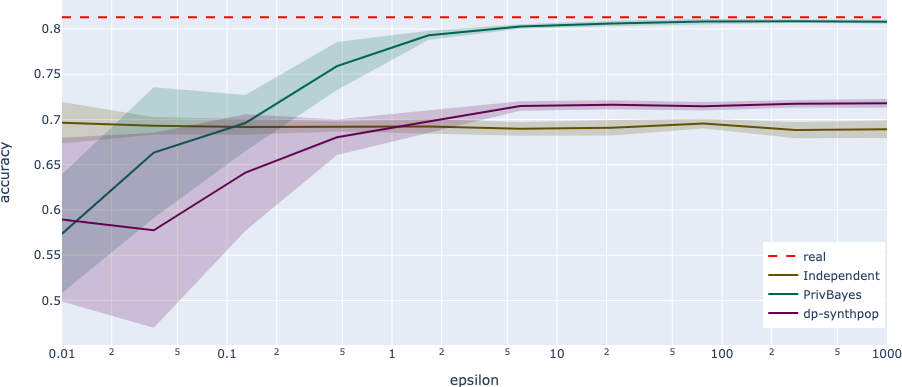}\label{fig:3b}}
	\caption{Impact of privacy on three specific instances of \emph{dpart} (\emph{Independent}, \emph{PrivBayes}, and \emph{dp-synthpop}), evaluated on simplified Adult.}
	\label{fig:3}
\end{figure*}

\descr{dp-synthpop.}

Yet another instance of our framework, alas not DP, is synthpop.
The original model expects either an explicit dependency graph or a visit order as input.
It also uses regression and classification methods in order to capture and preserve higher-order correlations and trends in the data.

We built on top of it and propose a DP version, called \emph{dp-synthpop}.
For the dependency step, we achieve DP by utilizing the automatic ``infer'' option in our framework, i.e., we iteratively select the column maximizing the mutual information with the already chosen columns in a noisy way using the Exponential mechanism.
As for estimating the conditional distributions, we rely on the DP predictive models from diffprivlib.
A possible dependency graph is visualized in Fig.~\ref{fig:2c} and a call to the model is presented below:

\begin{lstlisting}[language=Python,basicstyle=\tiny,frame=tb]
>>> from dpart.engines import DPsynthpop

>>> dpart_dpsp = DPsynthpop(epsilon, bounds=X_bounds).fit(X)
>>> synth_df = dpart_dpsp.generate(1000)
\end{lstlisting}

\section{Comparison}
\label{sec:4}

In order to assess the quality and performance of the specific instances presented in Sec.\ref{sec:3}, we run a quick experiment on a simplified version of the Adult dataset~\cite{dua2017adult} (we choose a subset of the columns).
The dataset comes with a binary classification task to predict whether the income of an individual is in excess of \$50k given some demographic information.

All models are trained with privacy budgets (or epsilon) ranging from 0.01 to 1,000 (where smaller value means tighter privacy guarantees).
For each epsilon value, the models are trained 5 times and for each trained model 5 synthetic datasets of size equivalent to the input are generated, leading to 25 datapoints per model per epsilon value.
We evaluate the resulting synthetic datasets from two standard angles: 1) mean marginal similarity (the average marginal distribution similarity across all columns between real and synthetic data) and 2) accuracy (accuracy of a decision tree classifier trained on real/synthetic dataset and evaluated against a held-out test dataset).
We demonstrate that even these simple baseline models could achieve competitive results.

\descr{Similarity.}

Overall, looking at the similarity in Fig.~\ref{fig:3a}, all models improve as a larger privacy budget is allocated.
Interestingly, \emph{Independent} outperforms the other two models for small privacy budgets (epsilon $<=1$) but this can be explained by the fact that it is particularly suited to capture marginal distributions and it does not ``waste'' its budget on irrelevant steps.
\emph{PrivBayes} seems to capture the marginals better than \emph{dp-synthpop} for all privacy budgets.

\descr{Accuracy.}

When it comes to accuracy, displayed in Fig.~\ref{fig:3b}, both \emph{PrivBayes} and \emph{dp-synthpop} are positively correlated with the increase of the privacy budget.
\emph{PrivBayes} approaches the real baseline for epsilon $>=1$.
\emph{dp-synthpop}'s underperformance compared to \emph{PrivBayes} is likely due to the use of DP linear and logistic regressions as underlying methods (while the default method behind the original synthpop is CART) as DP decision tree and DP random forest are significantly slower.
Unsurprisingly, \emph{Independent} does not perform well in this task.

\section{Conclusion}

In this work, we introduced \emph{dpart}, a general open source framework for DP synthetic data generation.
All readily available specific use cases of our framework (\emph{Independent}, \emph{PrivBayes}, and \emph{dp-synthpop}) are DP allowing for the simple and safe use of a varied set of generative approaches.
Furthermore, the \emph{dpart} interface provides great flexibility ensuring that more complex approaches or customized behaviors and domain knowledge can be easily configured.

The framework can be extended through the implementation of new DP methods, the introduction of new dependency building strategies as well as new readily available instances to ease the use of well-known and documented models.
We are looking forward to keep improving on this framework with the help of a growing community of researchers and practitioners.

\newpage

{\small\bibliography{mybib}}
\bibliographystyle{icml2022}

\end{document}